\documentclass[10pt,journal]{IEEEtran}
\usepackage[pdftex]{graphicx}
\DeclareGraphicsExtensions{.pdf,.jpeg,.png}

\usepackage{amsmath, amsfonts, amsbsy, amssymb,color}
\usepackage{amsmath,amssymb}
\usepackage{amsthm}
\usepackage{algorithm}
\usepackage{multirow, adjustbox}
\usepackage[noend]{algpseudocode}
\makeatletter
\def\BState{\State\hskip-\ALG@thistlm}
\makeatother

\usepackage{multirow,setspace,verbatim,amsfonts,graphicx,amsmath,amsthm,amsbsy,amssymb,epsfig,url}

\usepackage{epsfig}
\usepackage{graphicx}
\usepackage{amsmath}
\usepackage{amssymb}
\usepackage{color}
\usepackage{subfigure}

\newcommand{\kb}{{\mathbf k}}

\newcommand{\rb}{{\mathbf r}}

\newcommand{\Rd}{{\mathbb R}}

\newcommand{\Xc}{{\mathcal X}}
\newcommand{\Yc}{{\mathcal Y}}

\newcommand{\Qc}{{\mathcal Q}}
\newcommand{\Gc}{{\mathcal G}}

\newcommand{\Fc}{{\mathcal F}}

\newcommand{\Sc}{\mathcal{S}}

\newcommand{\Mc}{{\mathcal M}}
\newcommand{\Pc}{{\mathcal P}}

\usepackage[pagebackref=true,breaklinks=true,letterpaper=true,colorlinks,bookmarks=false]{hyperref}

\begin{document}

\title{Deep Residual Learning for Accelerated MRI using Magnitude and Phase Networks}
\author{Dongwook Lee, Jaejun Yoo, Sungho Tak and Jong Chul Ye, \IEEEmembership{Senior Member,~IEEE}
\thanks{Copyright (c) 2017 IEEE. Personal use of this material is permitted. However, permission to use this material for any
other purposes must be obtained from the IEEE by sending an email to pubs-permissions@ieee.org.}%
\thanks{DL, JY and JCY are with the Department of Bio and Brain Engineering, Korea Advanced Institute of Science and Technology (KAIST), Daejeon 34141, Republic of Korea (e-mail: jong.ye@kaist.ac.kr).}
\thanks{ST is with Bioimaging Research Team, Korea Basic Science Institute, 162 Yeongudanji-ro, Ochang-eup, Cheongwon-gu, Cheongju-si, Chungcheongbuk-do 28119, Republic of Korea}

}

\maketitle

\begin{abstract} 
~{\emph {Objective:}} Accelerated magnetic resonance (MR) scan acquisition  with compressed sensing (CS) and parallel imaging  is a powerful method to reduce MR imaging scan time. However, many reconstruction algorithms have high computational costs.  To address this, we investigate  deep residual learning networks to remove aliasing artifacts from artifact corrupted images. {\emph {Methods:}} The deep residual learning networks are composed of magnitude and phase networks  that are separately trained. If both phase and magnitude information are available, the proposed algorithm can work as an iterative  k-space interpolation algorithm using framelet representation.  When  only magnitude data is available, the proposed approach works as an image domain post-processing algorithm. {\emph {Results:}} Even with strong coherent aliasing artifacts, the proposed network successfully learned and removed the aliasing artifacts, whereas current parallel and CS reconstruction methods were unable to remove these artifacts. {\emph {Conclusion:}} Comparisons using single and multiple coil show that the proposed residual network provides good reconstruction results with orders of magnitude faster computational time than existing compressed sensing methods. {\emph {Significance:}} The proposed deep learning framework may have a great potential for accelerated MR reconstruction by generating accurate results immediately. 
\end{abstract}

\section{Introduction}

Magnetic resonance imaging (MRI) is one of the most valuable imaging methods for diagnostic and therapeutic indications. However, physical and physiological constraints limit the rate of magnetic resonance (MR) scan acquisition. One of the main MRI shortcomings is the long scan, hence accelerated MR acquisition is important to reduce scan time. Hence, k-space  under-sampling is often required, and  various reconstruction methods have been developed, including parallel MRI (pMRI)~\cite{pruessmann1999sense, griswold2002generalized} and compressed sensing (CS) MRI (CS-MRI)~\cite{donoho2006compressed,lustig2008compressed} to provide accurate reconstruction from  insufficient k-space samples.

Generalized auto-calibrating partial parallel acquisition (GRAPPA) \cite{griswold2002generalized} is a representative pMRI technique that interpolates the missing k-space data by exploiting the diversity of  coil sensitivity maps. CS-MRI reconstructs a high resolution image from randomly subsampled k-space data, utilizing data sparsity in the transformed domain.   These algorithms are usually combined to provide acceleration~\cite{HoYeKi06,jung2009k}.

Recent algorithms convert CS-MRI and/or pMRI problem to a k-space interpolation problem  using a low rank structured matrix completion, e.g. SAKE \cite{shin2013calibrationless} and the  annihilating filter-based low rank Hankel matrix approach (ALOHA)~\cite{jin2016general,lee2016acceleration,lee2016reference,ye2016compressive}, producing   state-of-the-art   reconstruction performance. However, the algorithms are limited by their computational complexity, and  CS algorithms require access to the k-space data, so the quality improvement in the image domain using post processing techniques is not possible.

Deep learning is an important framework for computer vision research \cite{krizhevsky2012imagenet} thanks to the availability of massive datasets and high performance computing hardware, such as graphic processor unit (GPU). In particular, convolutional neural networks (CNNs) train convolution filters to effectively extract  features from the training data, with great successes in classification~\cite{krizhevsky2012imagenet},  and regression problems, such as segmentation~\cite{ronneberger2015u}, denoising~\cite{mao2016image, zhang2016beyond}, super resolution~\cite{dong2014learning,kim2016accurate,ledig2016photo}, etc.  { Therefore, the current study utilized deep CNN to capture image structure features for medical image reconstruction.}

Wang et al~\cite{wang2016accelerating}  applied deep learning to CS-MRI, training the CNN from down-sampled reconstruction images to learn fully sampled reconstruction.  They then used the deep learning result  as either an initialization  or regularization term in classical CS approaches.  A multilayer perceptron was proposed for accelerated pMRI \cite{kwon2016learning}, and deep CNN architecture using an unfolded iterative CS algorithm was proposed in \cite{hammernik2016learning}, where trained optimal regularizers were proposed in contrast to the usual approach using handcrafted regularizers~\cite{hammernik2016learning}. Kang et al \cite{kang2016deep} provided the first systematic study of deep CNN for low dose x-ray computed tomography (CT), and showed that a deep CNN using directional wavelets was more efficient in removing low dose related CT noise. 

However, in contrast to low dose artifacts  from reduced tube currents, streaking originating from sparse projection views are globalized artifacts and difficult to remove using conventional denoising CNNs. Han et al \cite{han2016deep} and Jin et al \cite{jin2016deep} independently proposed multi-scale residual learning networks using U-Net \cite{ronneberger2015u} to remove these global streaking artifacts, and domain adaptation from CT to MRI has been also successfully demonstrated \cite{han2017deep}. These pioneering works show the significant potential of deep learning for image reconstruction.

The residual network (ResNet)~\cite{he2016deep} has been recently studied in the context of image classification. The gradient can be backpropagated into much deeper layers due to the identity mapping path in the residual network, which enables very deep residual network structure~\cite{he2016deep}. Therefore, we investigated ResNet based image reconstruction to deal with complex-valued MR data, and propose a new scheme to train phase and magnitude separately.  An important advantage of the proposed approach is that the algorithm does not require any k-space data for magnitude-only images, because the processing can be done in the image domain as an image restoration task. If both phase and magnitude information are available, the proposed algorithm works as an iterative  k-space interpolation algorithm using framelet representation~\cite{cai2008framelet,cai2009convergence}. Based on recent deep convolutional framelet theory, we employ a multi-scale network structure (called dual frame U-net)~\cite{han2017framing} to correct globalized artifacts from under-sampled k-space data. Numerical experiments confirmed the effectiveness of the proposed deep learning approach for CS and pMRI.

\section{Related Studies}

Consider a two-dimensional (2D) acquisition scenario where a 2D image $\alpha(\rb)$,~$\rb=(r_x,r_y)$ is reconstructed from a Fourier transform (FT) on k-space coordinates $\kb=(k_x,k_y)$. 
The k-space data $g_i(\kb)$ from the $i$-th coil can be represented as
\begin{equation}\label{eq:coil}
g_i(\kb) = \int \alpha(\rb) s_i(\rb) e^{-j2\pi \kb^T \rb} d\rb, \quad i=1,\cdots, N_{c},
\end{equation}
where $s_i(\rb)$ is the coil sensitivity of the $i$-th coil, and $N_{c}$ is the number of the coils. The  operator notation for \eqref{eq:coil} is 
$$g_i = \Fc [\Sc_i] \alpha,\quad i=1,\cdots, N_{c},$$
where $[\Sc_i]$ is a diagonal operator of the $i$-th coil sensitivity.

Only a subset of k-space measurements are obtained for accelerated MRI.  For a given subsampling pattern $\Lambda$, let $P_\Lambda$ be the diagonal matrix with 1 for indices in $\Lambda$ and 0 otherwise. Then, the standard reconstruction approach is to solve the penalized least squares minimization problem
\begin{eqnarray}\label{eq:prob}
\min_\alpha \sum_{i=1}^C \|y_i - P_\Lambda\Fc [\Sc_i] \alpha\|^2+\lambda R(\alpha),
\end{eqnarray}
where $R(\alpha)$ denotes regularization terms, such as the total variation penalty, etc. When the sensitivity map  $[\Sc_i]$ is unknown, direct estimation of sensitivity weighted images can be expressed as
\begin{eqnarray}\label{eq:zi}
x_i =  [\Sc_i] \alpha,\quad i=1,\cdots, N_{c}.
\end{eqnarray}
For example, $l_1$- iterative self-consistent parallel imaging reconstruction ($l_1$-SPIRiT)  \cite{lustig2010spirit} utilizes additional GRAPPA type constraints for the CS problem
\begin{eqnarray*}
\min_{\Xc} &&\|\Psi \Xc \|_{1,2} \\
\mbox{subject to}&& \Gc= P_\Lambda\Fc \Xc\\
&& \Xc=\Mc  \Xc,
\end{eqnarray*}
where 
\begin{eqnarray}\label{eq:def}
\Xc=\begin{bmatrix}x_1 & x_2 & \cdots &x_{N_{c}}\end{bmatrix}, ~ \quad \Gc=\begin{bmatrix}g_1 & g_2 & \cdots &g_{N_{c}} \end{bmatrix}
\end{eqnarray}
denote the unknown images and their k-space measurements for the given set of coils, respectively; $\|\cdot\|_{1,2}$ denotes the $(1,2)$-mixed norm of a matrix; $\Psi$ denotes a discrete wavelet transform matrix; and $\Mc$ is an image domain GRAPPA operator. This problem can be solved using proximal algorithms.

\begin{figure*}[!t] 
\vspace{-0.5cm}
  \centering
  \centerline{\includegraphics[width = 15cm]{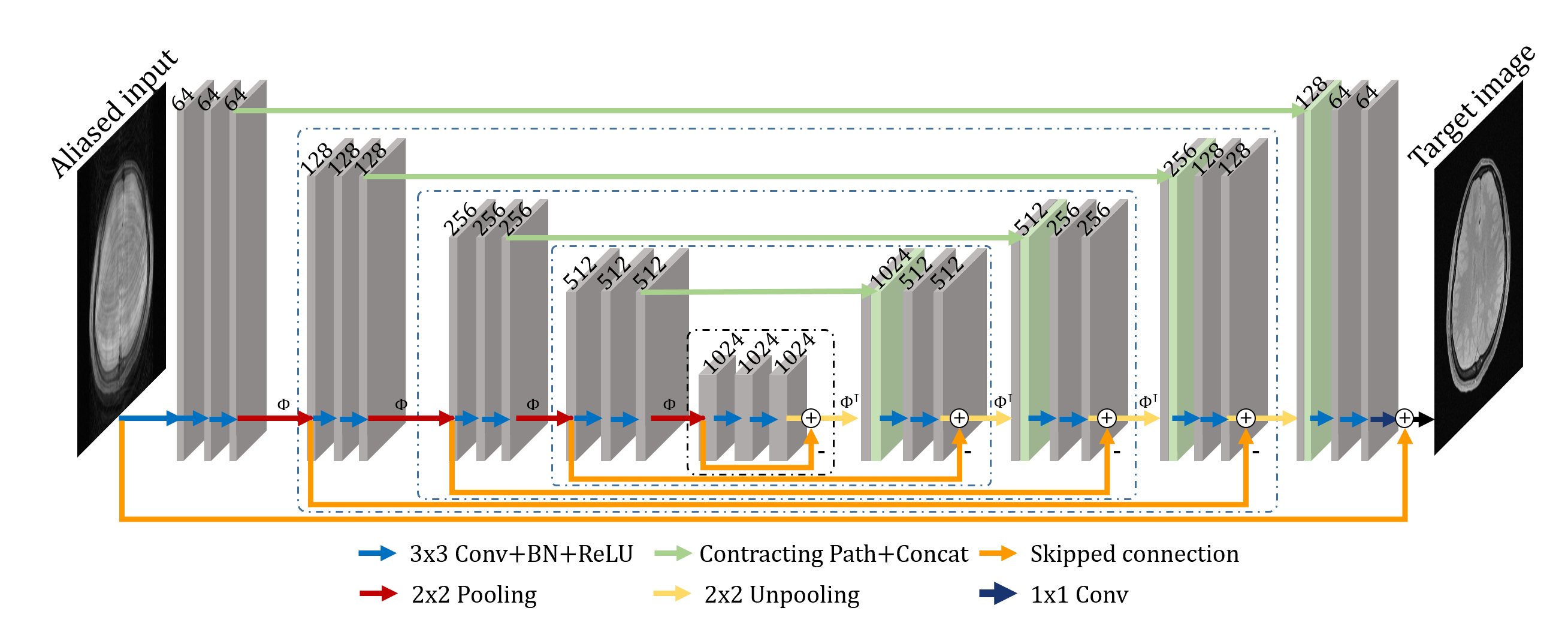}}
  \centerline{}\medskip
\vspace{-0.5cm}
\caption{Proposed multi-scale residual learning using dual frame U-net}\label{fig:network} 
\end{figure*}

\section{Main Contributions}\label{sec:MainContributions}

\subsection{Magnitude and Phase Networks}

We are interested in solving accelerated MR reconstruction problems using k-space interpolation. The concept was inspired by classical frame based inpainting approaches~\cite{cai2008framelet,cai2009convergence}.
Let $\Qc$ be the mapping that satisfy perfect reconstruction (PR), i.e.,
\begin{eqnarray}\label{eq:pr}
\Qc(\Yc) = \Yc , 
\end{eqnarray}
where $\Yc$ is the fully sampled k-space samples from multiple coils.
Then, we have the identity
\begin{eqnarray}
\Yc 
&=& \mu P_\Lambda \Gc + (I-\mu P_\Lambda) \Qc(\Yc),\quad 0\leq \mu\leq 1 ,  \label{eq:update0}
\end{eqnarray}
where 
$\Gc$ denotes that the measurement in \eqref{eq:def} are measured on $\Lambda$. To avoid the trivial solution, mapping $\Qc$ should include the shrinkage operator.
Equation~\eqref{eq:update0} can be expressed as an iterative $k$-space interpolation algorithm
\begin{eqnarray}\label{eq:update2}
\Yc_{n+1} = \mu P_\Lambda \Gc + (I- \mu P_\Lambda) \Qc(\Yc_n)  .
\end{eqnarray}
Since MRI uses FT, $\Yc = \Fc \Xc$, \eqref{eq:pr} can be expressed as
\begin{eqnarray}\label{eq:imagePR}
\Fc\Xc = \Qc(\Fc\Xc) \Longleftrightarrow & \Xc = \Fc^{-1}\Qc( \Fc \Xc) = \tilde\Qc(\Xc),
\end{eqnarray}
corresponding to PR image domain. 
Therefore, we want to find the image domain mapping $\tilde\Qc$.
The reason  we prefer  $\tilde \Qc$ in the image domain is that image domain networks have been widely investigated in the context of computer vision, whereas to the best of our knowledge, k-space domain networks has not be previously investigated.

Here, we must consider the complex nature of MR images to impose constraint \eqref{eq:imagePR}. 
Hence, we represent mapping \eqref{eq:imagePR} as
\begin{eqnarray}\label{eq:mpPR}
\Xc = \tilde \Qc(\Xc) = \Qc_\Mc(|\Xc|)\exp\left(j \Qc_\Pc(\angle \Xc) \right),
\end{eqnarray}
where $\Qc_\Mc$ and $\Qc_\Pc$ denote magnitude and phase mapping, respectively.
%
This approach has many advantages. Magnitude-only images are used for many clinical applications, and users process data from an image workstation  that does not have access to the raw k-space data. Hence, the magnitude network works as a post-processing algorithm to improve the image. Furthermore, many public datasets, such as the Human Connectome Project (HCP) dataset (https://db.humanconnectome.org), are in the form of magnitude images, which can be used to pre-train the magnitude network.

On the other hand, when phase information is available due to the access to the complex images, then combining the phase domain network allows exploiting iterative k-space interpolation using \eqref{eq:update2}, which  is  summarized in Algorithm~\ref{KMmethod}. Algorithm~\ref{KMmethod} is a relaxed iteration using the Krasnoselskii-Mann (KM) method~\cite{bauschke2011convex} to improve frame based interpolation iteration convergence.

\begin{algorithm}
\caption{ Krasnoselskii-Mann iteration for k-space interpolation}\label{KMmethod}
\begin{algorithmic}[1]
\State Train the image domain mapping $\tilde \Qc$ in \eqref{eq:mpPR} using the training dataset.
\State Set $0<\mu\leq1$ and $0<\lambda_n <1, \forall n.$
\State Initialize $\Yc_0=\Gc$.
\For {$n=1,2,\dots$ until convergence}
\State  $\Qc_n =\Fc\tilde\Qc\left(\Fc^{-1}\Yc_n\right)$
\State ${\bar\Yc}_{n+1} =\mu P_{\Lambda}\Gc+(I-\mu P_{\Lambda}) \Qc_n$
\State $\Yc_{n+1} =\Yc_n+\lambda_n\left(\bar{\Yc}_{n+1}-\Yc_n\right)$
\EndFor 
\end{algorithmic}
\end{algorithm}

Therefore, the remaining issue is how to design the magnitude and phase networks, $\Qc_\Mc$ and $\Qc_\Pc$, respectively. This is discussed in Section~\ref{sec:MainContributions} C.


\subsection{Network Architecture}

\begin{figure}[!b]
    \centerline{\includegraphics[width=7.5cm]{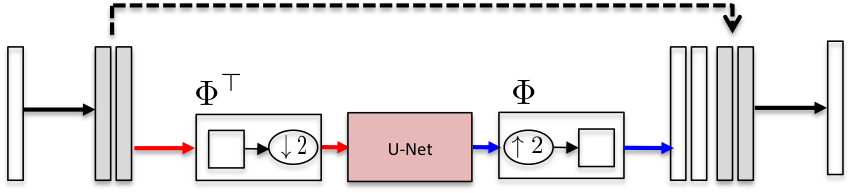}}
    \centerline{\mbox{(a)}}
        \centerline{\includegraphics[width=7.5cm]{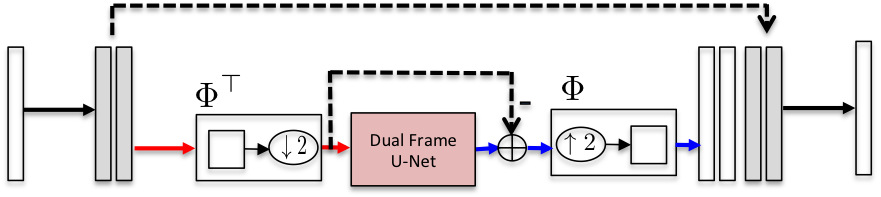}}
    \centerline{\mbox{(b)}}
    \caption{(a) Standard U-net, and (b) dual frame U-net. Dashed lines refer to the skipped connections, square box within $\Phi,\Phi^\top$  corresponds to average pooling and unpooling, respectively. Next level units are recursively added to the low frequency band signals.}
    \label{fig:Unet}
\end{figure}

One of the most widely used deep learning network architecture for inverse problems is U-net~\cite{jin2016deep,han2017deep}, which incorporates bypass connection and pooling/unpooling layers. Figure~\ref{fig:Unet}(a) shows an example one-dimensional (1D) U-Net.  However, U-net has limitations due to duplication of low frequency content~\cite{ye2017deep}.  Therefore, we employ the dual frame U-net~\cite{han2017framing}, as shown in Fig.~\ref{fig:Unet}(b) to overcomes this limitation. In contrast to U-net, the residual signal at low resolution is up-sampled through the unpooling layer and subtracted to remove duplicated low frequency content. This can be implemented using an additional bypass connection for the low resolution signal, as shown in Fig.~\ref{fig:Unet}(b).   See Appendix~\ref{ap:dual_unet} and  \cite{han2017framing} for detailed descriptions regarding why and how to employ dual frame U-Net. In addition to the application of dual frame U-Net for  sparse CT reconstruction problems~\cite{han2017framing}, dual frame U-Net has been successfully used for  elastic source imaging problems~\cite{yoo2018mathematical}, a complicated inverse scattering problem, confirming the advantages of dual frame U-Net over U-Net. Therefore, we propose dual frame U-Net for MR reconstruction.

Figure \ref{fig:network} shows the 2D extension of Fig. \ref{fig:Unet}(b). Since there is a skipped connection at the output level, this is referred to as a residual network. The proposed residual network consists of a convolution layer, batch normalization~\cite{ioffe2015batch}, rectified linear unit (ReLU)~\cite{krizhevsky2012imagenet}, and contracting path connection with concatenation~\cite{ronneberger2015u}. Each stage contains four sequential layers comprising convolution with $3\times3$ kernels, batch normalization, and ReLU layers.  The final layer only contains a convolution with $1\times1$ kernel. Figure \ref{fig:network} shows the number of channels for each convolution layer. Using pooling layers, the proposed network has five representation scales, as shown in Fig.~\ref{fig:network},  where  the number of channels after each pooling layer is doubled.  Input resolution is $256 \times 256$, and the representation resolution is halved for each scale change until it becomes $16 \times 16$ in scale 4.  The pooling layers mean the receptive field in the multi-scale network increases  exponentially. Consequently, the dual frame U-net structure produces a larger receptive field, which is useful to capture global artifact patterns.

\subsection{Reconstruction Flow }

{The networks for the reconstruction of magnitude and phase have the same architecture as shown in Fig.~\ref{fig:network}.} To train the network, we first apply the inverse Fourier transform to the down-sampled k-space data to generate aliased images. Inputs for the magnitude network were the magnitude of distorted MR images and targets were the alias-free image magnitudes. Phase reconstruction was similarly trained with distorted and alias-free image phases, respectively.

\begin{figure}[!hbt]    
\centering
\center{\includegraphics[width=8cm]{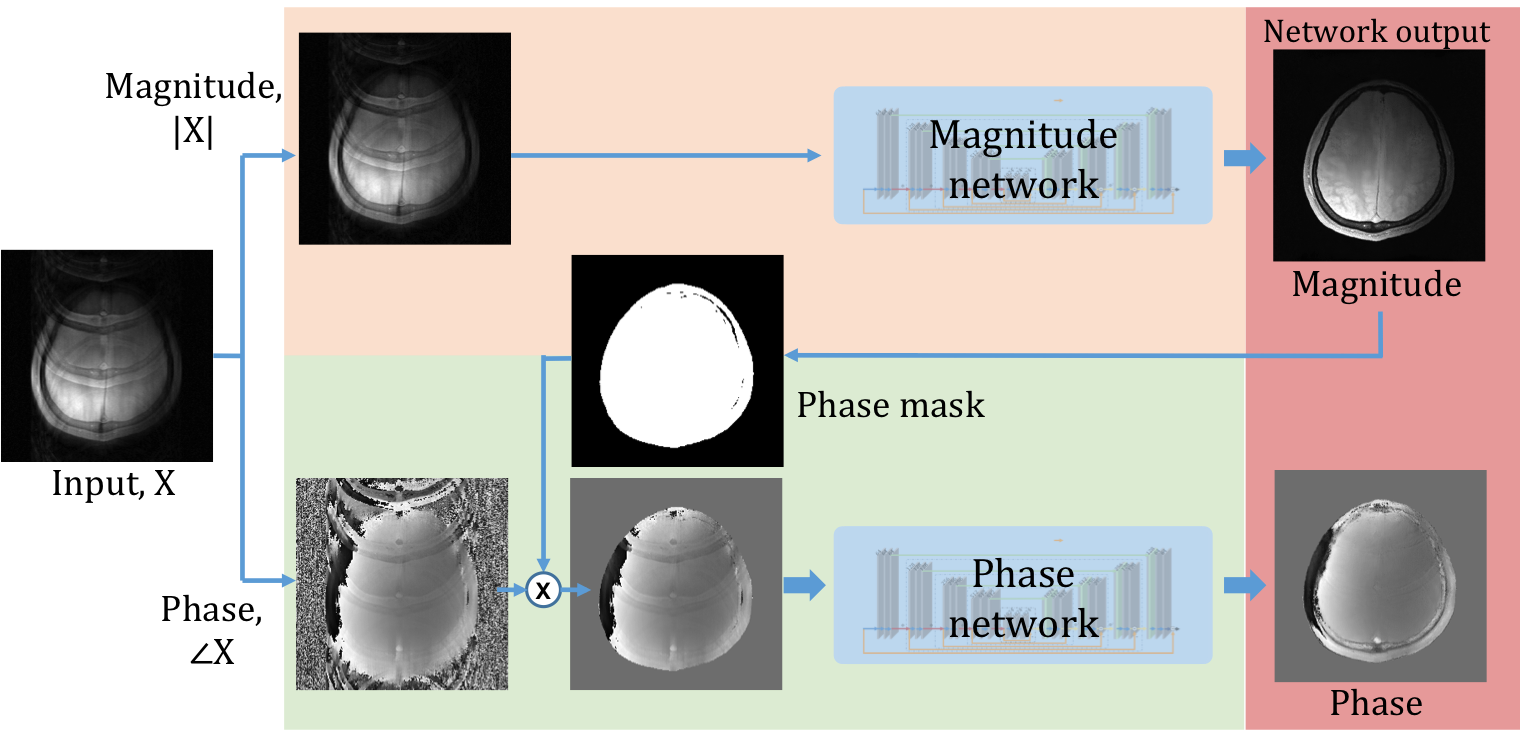}}
\caption{Magnitude and phase image reconstruction}
\label{fig:recon_flow}
\end{figure}

Although both networks have the same structure,  there is an additional step for the phase network.  The phase image of Figure~\ref{fig:recon_flow} (bottom row) shows that the phase image region within the brain has smooth structures with slowly varying pixel values from $-\pi$ to $+\pi$, whereas the region outside the brain has approximately zero magnitude, but large phase fluctuations, since the pixels have random values from $-\pi$ to $+\pi$. These random phases outside the brain region  complicate network training.   One advantage of separate phase estimation is that the spatial support information from the magnitude network can be used to improve phase reconstruction. Therefore, we use phase masking to eliminate these random phase effects outside the brain. 

Thus, we first trained the magnitude network, and then obtained phase masks from reconstructed magnitude images using a simple threshold. We set phase outside the ROI (as defined by the phase masks) to zero for both input and artifact phase images, eliminating random phase effects outside the brain. The phase network is then trained using artifact-free phase data within the phase mask as targets.

After training the networks, {the inference of the networks follows the same steps as the training process }(Fig.~\ref{fig:recon_flow}). Since the phase masks should be generated before phase reconstruction, we first reconstruct magnitude images using the trained magnitude network (Fig.~\ref{fig:recon_flow}, orange box).
The phase mask indicating the brain area is then generated by thresholding the reconstructed magnitude image, and phase outside the phase mask is removed before being applied to the phase network (Fig.~\ref{fig:recon_flow}, green box). For the multi-channel case, the square root of sum of squares (SSOS) image was calculated after reconstruction.

If phase information is available, we can use Algorithm~\ref{KMmethod} to improve reconstruction. Due to the recursive nature of the network, the following iterative training procedure was used. The magnitude and phase network are trained on the initial training database, and missing k-space measurements are estimated. We then augment measured k-space data with estimated k-space data, and obtain reconstructed images by FT. These images are used as an additional dataset for further magnitude and phase network training. Thus, the networks learn the mapping not only from strongly aliased input, but also from almost artifact-free images. {For the inference of the networks,} $\lambda_n$  in Algorithm~\ref{KMmethod} is set to decrease from 1 to 0,  and $\mu$ is set to 1.

The theoretical background for the proposed training using both strongly aliased and near artifact-free images is from the framelet nature of deep CNNs, which was the main outcome from the mathematical theory in our companion paper~\cite{ye2017deep} (see Appendix A for the review). Neural network training learns framelet bases from the training data that best represent the signals. Thus, the learned bases should be sufficiently robust to provide good representation for the given input data set. In the KM iteration, each iterative step provides improved images, which should be input to the same neural network. Thus, framelet bases should be trained to be optimal not only for strongly aliased input but also for near artifact-free images. Similar training techniques are implicitly used for recursive neural networks (RNN) after {unfolding~\cite{werbos1990backpropagation}.}

\begin{figure*}[!t]
\vspace{-0.5cm}
  \centering
  \centerline{\includegraphics[width = 17cm,height=10cm]{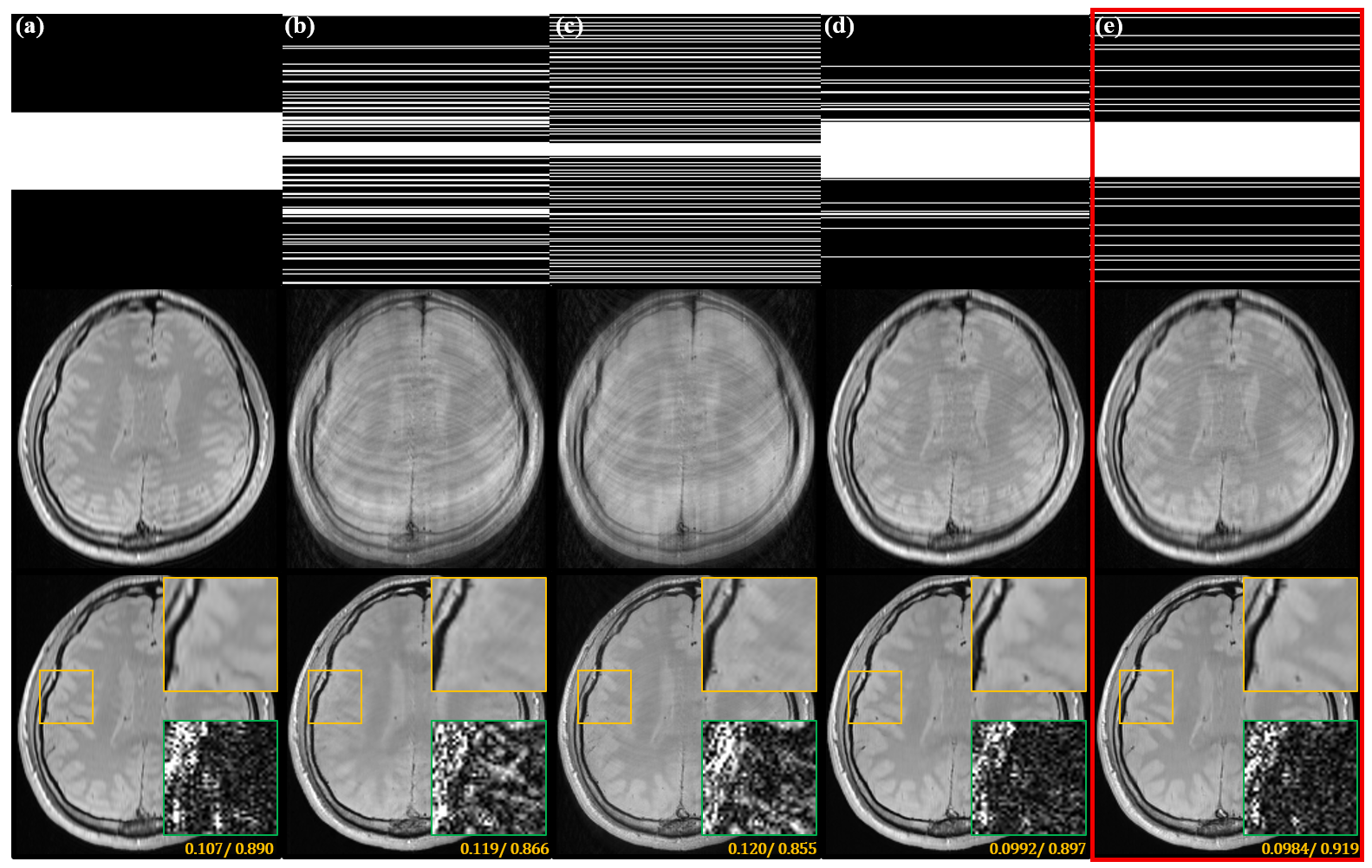}}
  \centerline{}\medskip
\vspace{-0.5cm}
\caption{Sampling pattern effect for the proposed deep residual learning approach.  Sampling patterns: (a) Low frequency only; (b) and (c) Gaussian random and uniform random sampling with twelve ACS lines (5\% of PE), respectively; (d) and (e) are Gaussian random and uniform random sampling with 52 ACS lines (20\% of PE), respectively; Net acceleration rate $\approx \times 3.5$.  Row 1: sampling patterns; Row 2: corresponding aliased inputs; Row 3: corresponding reconstructions. Magnified and corresponding error images are shown.}\label{fig:SamplingRecon} 
\end{figure*}

\section{Method}

\subsection{Magnetic Resonance Dataset}

We used three brain image datasets, including structural MRI images from the HCP dataset, to increase the magnitude network training dataset; and two {\emph {in vivo}} k-space datasets from 3T and 7T scanners. We select 200 of the 1200 HCP T2 weighted structural images to pre-train the magnitude network. The data set were collected using the 3D T2-SPACE sequence with TR $=$ 3200~ms, TE $=$ 565~ms, variable flip angle, FOV $=$ 224$\times$224~mm, and 32 channel head coils. A total of 12,000 slices of the HCP dataset were used for pre-training.  Since the HCP T2 weighted image was obtained as the SSOS from the multi-coil reconstruction, we generated simulated coil magnitude images by multiplying the sensitivity maps  extracted from 3T and 7T {\emph {in vivo}} MR datasets.

Although public dataset such as HCP can be used to pre-train the magnitude network,  phase variations in MR images are usually small, and a small number of {\emph {in vivo}} data are usually sufficient to train the phase network. Therefore, the phase network was not pre-trained. 

The {\emph {in vivo}} data 3T MR dataset  comprised 81 axial brain slices from nine subjects. Images from seven subjects were used for training (61 images),  one subject for validation (8 images), and  the final subject (12 images) for testing. Data were acquired in Cartesian coordinates from a 3T MR scanner with four Rx coils (Siemens, Verio). Spin echo (SE) and gradient echo (GRE) scans used TR $=$ 3000--4000~ms, TE $=$ 4—20 ms, FA $= 90^\circ$, slice thickness $=$ 5 mm, 256$\times$256 acquisition matrix, four coils, FOV $=$ 240$\times$240 mm, and anterior to posterior PE direction. Brain images in the dataset had different intensity and maximum scales, since they were acquired with various scan conditions (GRE/SE, various TE/TR, etc.). 
We  selected the first coil data (of four) for single-channel experiments, and used all the coil images for parallel imaging experiments.

\subsection{Down-Sampled Data Generation}

The original k-spaces were retrospectively down-sampled, according to several  under-sampling patterns. We compared five sampling patterns to identify any performance effects that differed regarding the ACS lines and distribution type (Gaussian and random), as shown  in Fig.~\ref{fig:SamplingRecon} (top row). Additional low frequency auto-calibration signal (ACS) lines were required to ensure that aliasing artifacts were primary high frequency signals rather than low frequency repetitions.  ACS lines were also necessary for GRAPPA reconstruction for comparative studies.  The acceleration factor was set to 3.5 for all sampling patterns.

Normalization is an important preprocessing step to improve reconstruction performance. Since the images were acquired under various scan conditions, the data was normalized separately for each image. Input data for the magnitude network were normalized using scaling factor $a = 256/\|x\|_\infty$, where $\|x\|_\infty$ is the maximum image pixel intensity, ensuring input image magnitude was limited to [0,256]. Phase images were already normalized from $-\pi$ to $\pi$, but we modified the range to [0,2$\pi$] for better reconstruction performance.

Data augmentation was essential for the {\emph {in vivo}} data set to make the network more robust.  Hence, the original full-sampled images were transformed by flipping to produce augmented MR images. The transforms were performed on the complex image domain to retain full and down-sampled k-space data for the augmented MR images.
Thus, 4 more training samples were generated from {\emph {in vivo}} k-space data for augmentation.

\subsection{Network Training }

The network was implemented in MATLAB 2015a using the MatConvNet toolbox (version~23)~\cite{vedaldi2015matconvnet} (Mathworks, Natick). We used a GTX 1080 graphic processor and i7-4770 CPU (3.40 GHz).  The convolutional layer weights were initialized by Gaussian random distribution using Xavier's method \cite{glorot2010understanding} to obtain an appropriate proper scale. This avoided the signal exploding or vanishing in the initial training phase. We chose the $l_2$ loss function and stochastic gradient descent (SGD) method with momentum to train the weights by minimizing the loss function. Early stopping was used as regularization to avoid overfitting.

\begin{figure}[!h] 
\center{\includegraphics[width=9cm]{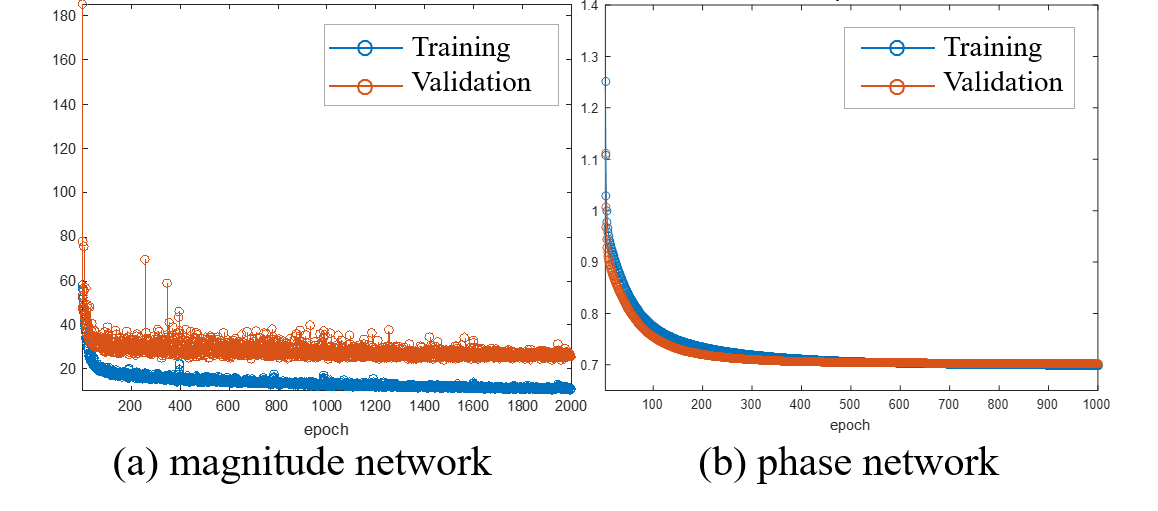}}
\vspace*{-0.5cm}    
\caption{Convergence for (a) magnitude and (b) phase network training.  (Blue) training set, and (orange) validation set. Uniform random sampling with 20\% ACS lines as shown in Fig.~\ref{fig:SamplingRecon}(e). }
\label{fig:train_curve}
\end{figure}

To use the public HCP data set, the training step was divided  into three steps:  the first 90 epochs for pre-training using the HCP data set, the next 1410 epochs for 3T or 7T {\emph{ in vivo}} data, and the final 500 epochs for recursive training. Learning rate was reduced logarithmically from  $10^{-1}$ to $10^{-2}$ each epoch. The number of epochs for phase network  training was set to 1000 due to the relatively simpler structures of phase compared to magnitude, and mini-batch size was set to 10. Figure~\ref{fig:train_curve} shows the phase and magnitude network training convergence curves.
Training each network for the 3T and 7T MR datasets took approximately 14 and 20 hours, respectively, including pre-training and iterative training using the reconstructed images for KM iteration.  Pre-training took approximately 7--8 hours for the HCP dataset.

\subsection{Comparative Studies}

We used the ALOHA~\cite{jin2016general} reconstruction as a representative  CS algorithm for both single and multi-channel reconstruction to verify network performance. We also compared the reconstruction results for multi-channel dataset with those of GRAPPA~\cite{griswold2002generalized}.  Since GRAPPA is designed for uniform sampling, the GRAPPA reconstruction was performed for uniform sampled k-space with the same ACS lines and same acceleration factor ($\times 3.5$). The $5 \times 5$ kernel was chosen for the GRAPPA reconstruction, and the following parameters were used for the ALOHA: filter size $=11 \times 11$, $\mu = 1$, two-stage pyramidal decomposition, tolerance set $=$ [0.1, 0.01], and spline wavelet for k-space weighting.

We used SSOS on output magnitude images for final reconstruction as the objective comparison,  and SSOS of alias-free  reconstruction as the ground-truth, 
$$\hat x_{SSOS} = \sqrt{\sum_{i}^{N_c} |\hat x_i|^2}, $$
where $\hat x_i$ refers to the estimated $i$-th coil image, and $N_C$ denotes  the number of coils. Phase images reconstructed from the full k-space data were used as phase space ground-truth.

Reconstruction performance was measured by the normalized root mean square error, 
$$ NRMSE =  \frac{ \left( \|\hat x - x \|_F^2/N \right)^{\frac{1}{2}} }{\bar x},$$
where $\hat x$ and $x$ refer to the estimate and ground-truth images, respectively;  $N$ denotes the number of pixels; and $\bar x$ is the mean value of $x$; and the structural similarity index~\cite{wang2004image}
$$SSIM = l^\alpha \cdot c^\beta \cdot s^\gamma ,$$ 
where $\alpha=\beta=\gamma=1$; and
$$l = \frac{2\mu_x \mu_{\hat{x}}+C_1}{\mu_x^2+\mu_{\hat{x}}^2+C_1},$$
$$c = \frac{2\sigma_x \sigma_{\hat{x}}+C_2}{\sigma_x^2+\sigma_{\hat{x}}^2+C_2},$$
and 
$$ s=\frac{\sigma_{x{\hat{x}}}+C_3}{\sigma_x\sigma_{\hat{x}}+C_3}$$  
are the luminance, contrast, and structural terms, respectively, where $\mu_x, \mu_{\hat{x}}, \sigma_x, \sigma_{\hat{x}}, \sigma_{x{\hat{x}}}$ are the local mean, standard deviation, and cross-covariance for images $x$ and ${\hat{x}}$, and $C_1, C_2$ and $C_3$ are predefined constants~\cite{wang2004image}.

\begin{figure}[!hbt]    
\center{\includegraphics[width=7cm]{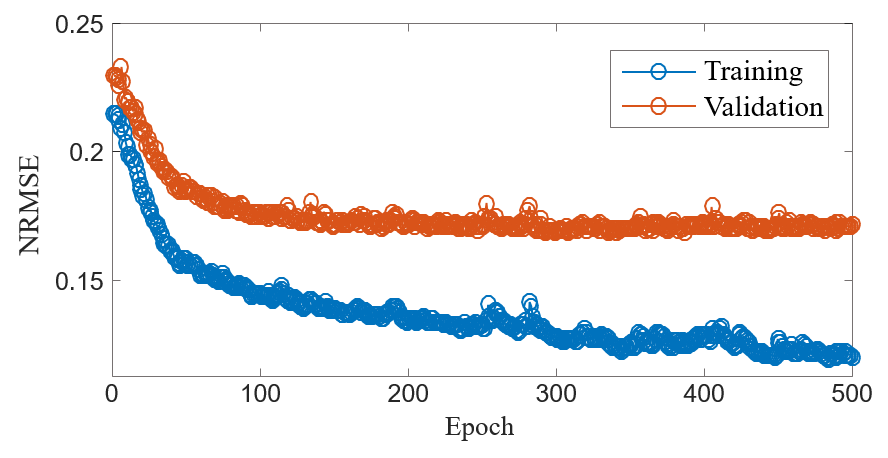}}
\caption{Training and validation phase convergence for a baseline network using real and imaginary channels.  Uniform random sampling with 20\% ACS lines as shown in Fig.~\ref{fig:SamplingRecon}(e).}
\label{fig:baseline}
\end{figure}

\section{Experimental Results}\label{sec:result}

We first investigated  reconstruction performance depending on various sampling patterns.
Figure~\ref{fig:SamplingRecon} shows the corresponding aliased input and final reconstruction results (center and bottom rows, respectively). The 20\% ACS lines were better than reconstruction results using  5\% ACS lines and low frequency  sampling, with uniform slightly better than Gaussian random sampling. Therefore, we employed uniform random sampling with 20\% ACS lines for all subsequent experiments.

\begin{figure*}[!hbt]   
\vspace{-0.5cm}
\centering
 \centerline{\includegraphics[width=15cm,height=8cm]{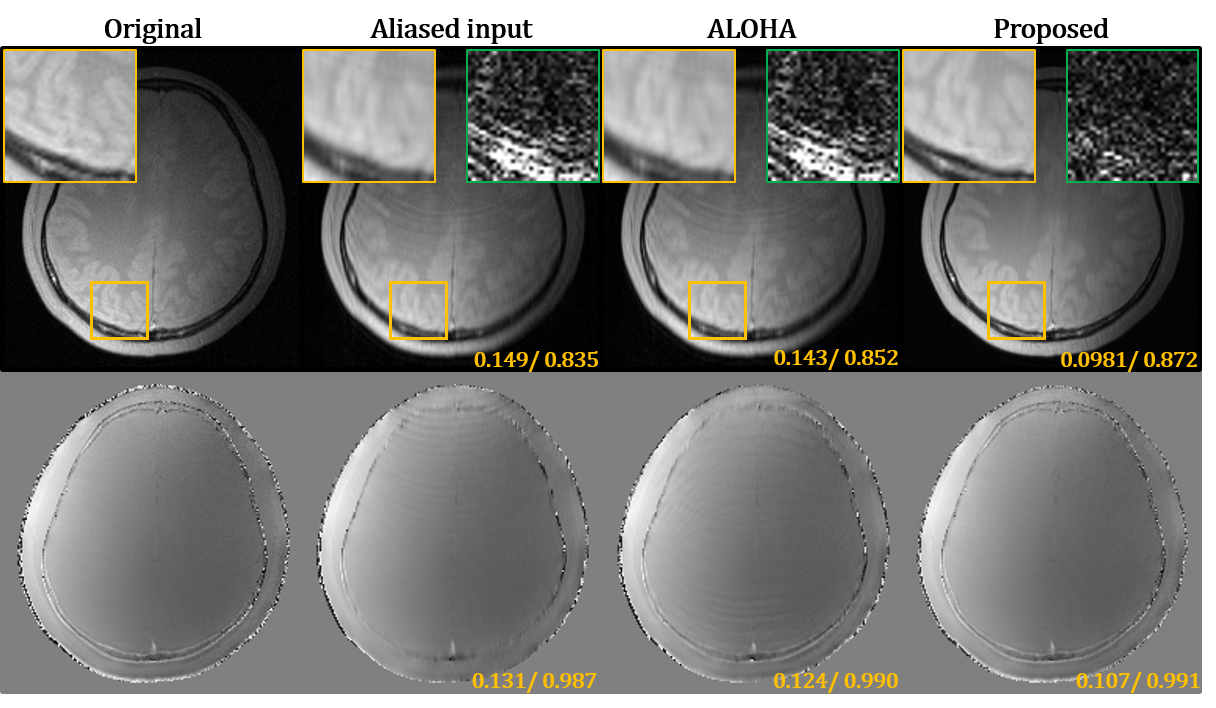}}
\vspace{-0.2cm}    
\centerline{\mbox{(a)}}
    \centerline{\includegraphics[width=15cm, height=4.5cm]{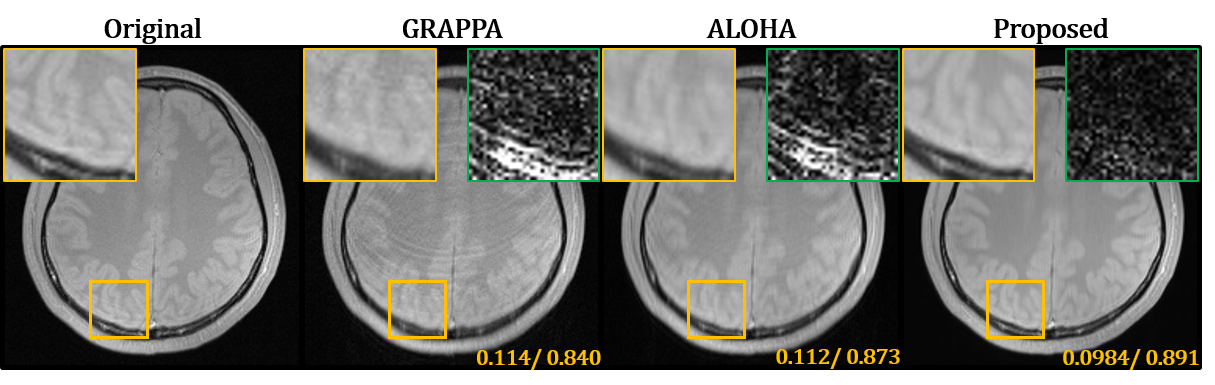}}
        \vspace{-0.2cm}
    \centerline{\mbox{(b)}}
\caption{Reconstruction results (row 1 is magnitude and row 2 phase image) for (a) single channel by ALOHA and proposed method; and (b) Multi-channel for magnitude images by GRAPPA, ALOHA, and proposed method. The square root of sum of squares (SSOS) was used for multi-channel MR images. Magnified and corresponding error images are shown. Uniform random sampling with 20\% ACS lines as shown in Fig.~\ref{fig:SamplingRecon}(e). }
\label{fig:CompResult}
\end{figure*}

We compared a baseline approach to handle the complex data and verify that separate magnitude and phase network training did not degrade performance. The baseline approach stacked the real and imaginary values as two-channel inputs to the neural network and learned the two-channel output consisting of real and imaginary images. After the reconstruction of real and imaginary images, complex valued images were obtained by combining real and imaginary valued images. The inputs were normalized in the same manner, with real and imaginary values [-256,256]. The real and image parts be negative, in contrast to the magnitude network that has only non-negative values. The baseline network architecture was exactly the same as the proposed network (Fig.~\ref{fig:network}), except that the input and labels were two-channels of real and image components of down-sampled and fully sampled reconstruction, respectively. The network was trained using exactly the same training dataset and procedure, and Fig.~\ref{fig:baseline} shows the convergence.

\begin{table}[!b]
\centering
\caption{Comparison between the proposed and complex networks by the normalized root mean square error (NRMSE) and the structural similarity index (SSIM)
}
\label{cmpComplexNet}
\begin{adjustbox}{width=0.4\textwidth}
\begin{tabular}{ccccc}
\hline
\multirow{2}{*}{} & \multicolumn{2}{c}{Magnitude} & \multicolumn{2}{c}{Phase} \\ \cline{2-5} 
                  & NRMSE         & SSIM          & NRMSE       & SSIM        \\ \hline 
Input             & 0.237         & 0.867         & 0.136       & 0.986       \\
Complex-net       & 0.154         & 0.903         & 0.111       & 0.989       \\
Proposed          & 0.155         & 0.914         & 0.111       & 0.990       \\ \hline
\end{tabular}
\end{adjustbox}
\end{table}

Table~\ref{cmpComplexNet} shows that the baseline and proposed networks have similar performance, although the proposed network has slight better SSIM. Therefore, considering the advantages of separate magnitude and phase image training, we used the proposed magnitude and phase network structures for subsequent study. 

Performance of the proposed network performance was  compared with that of ALOHA and GRAPPA to demonstrate the competitive edge against CS and parallel imaging approaches, as shown in Fig.~\ref{fig:CompResult}. Fig.~\ref{fig:CompResult}(a) shows that single channel images had significant aliasing artifacts in the input image. However, the aliasing artefacts were mainly for edge images due to the additional ACS lines.
The ALOHA approach was somewhat successful with some remaining aliasing artifacts, and the ALOHA reconstruction image was better than the input. However, the result still has aliasing artifacts. In contrast, the approach produced accurate reconstruction, removing aliasing artifacts and has minimum NRMSE. 

Fig.~\ref{fig:CompResult}(b) shows parallel imaging experiments with four channel data. GRAPPA reconstruction exhibits many reconstruction errors, and although ALOHA reconstruction removed many aliasing artifacts, the results were not perfect. In contrast, the proposed method provided excellent reconstruction.

\begin{figure}[!hbt]    
\center{\includegraphics[width=8cm]{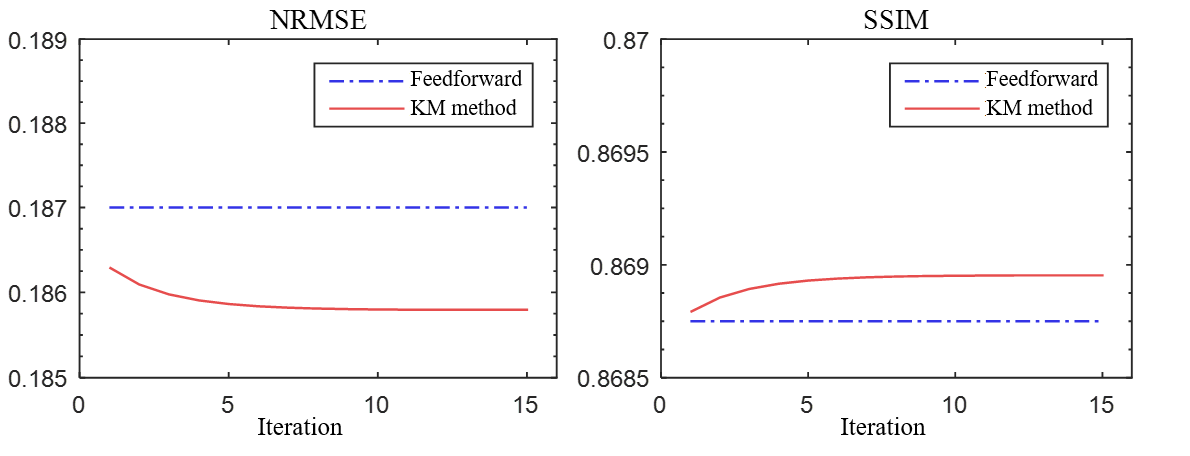}}
\caption{KM iteration in Algorithm~\ref{KMmethod} convergence. NRMSE and PSNR for magnitude reconstructions are also shown. Blue dashed lines represent feed-forward image network without KM iteration.  Uniform random sampling with 20\% ACS lines as shown in Fig.~\ref{fig:SamplingRecon}(e).
 }
\label{fig:iterRecon}
\end{figure}

Table \ref{my-label} summarizes performance metrics for pre-training using HCP data and iterative update using KM iteration. Pre-training using large datasets improves network reconstruction performance. KM iteration provided minimum NRMSE and maximum SSIM, as shown in Fig.~\ref{fig:iterRecon}.

\begin{table}[!hbt]
\centering
\caption{Performance for different training strategies. HCP refers to the human connectome datasets. }
\label{my-label}
\begin{adjustbox}{width=0.4\textwidth}
\begin{tabular}{llll}
\hline
No. &  Training method               & NRMSE & SSIM  \\ \hline
(A) & No HCP data pre-training   & 0.191 & 0.861 \\
(B) & HCP data pre-training   & 0.187 & 0.869 \\
\hline
\end{tabular}
\end{adjustbox}
\end{table}

\begin{figure}[!hbt]    
        \vspace{-0.5cm}
\center{\includegraphics[width=6cm,height=4cm]{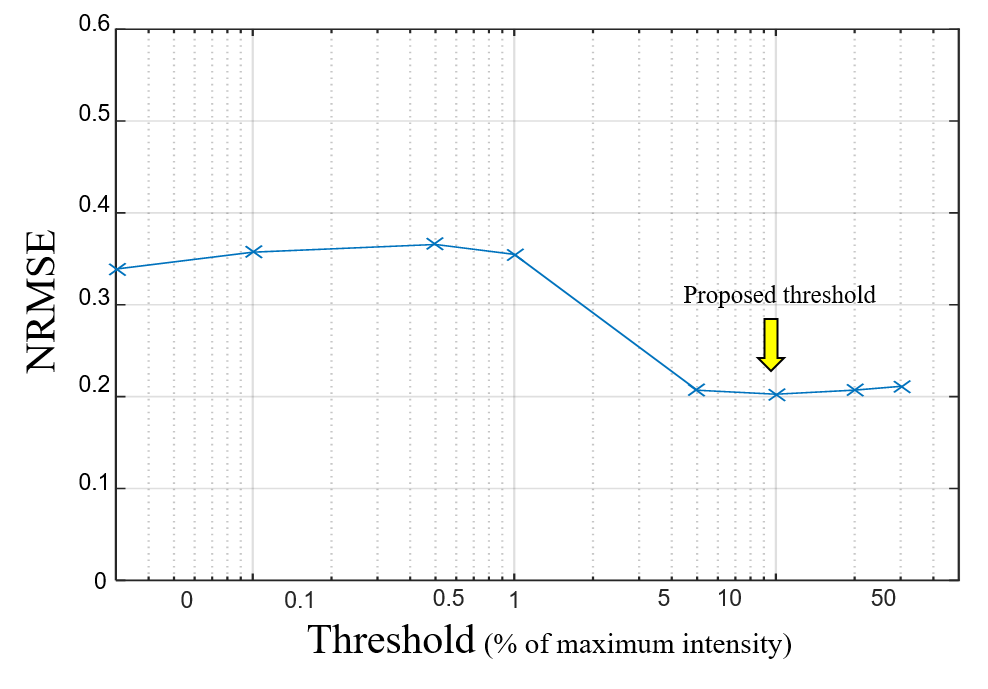}}
        \vspace{-0.3cm}
\caption{Phase mask threshold effects on phase reconstruction error (NRMSE). Uniform random sampling with 20\% ACS lines as shown in Fig.~\ref{fig:SamplingRecon}(e).}
\label{fig:phaseThresh}
\end{figure}

Since the proposed phase network requires a phase mask by thresholding the magnitude network, phase reconstruction accuracy may be sensitive to the chosen threshold. Therefore, the phase network was trained under various phase masking levels: 0, 0.1, 0.5, 1, 5, 10, 20, 30 \% of maximum intensity, where 0 means the phase network was trained without phase masking, and phase reconstruction NRMSE calculated, as shown in Fig.~\ref{fig:phaseThresh}. The network was not well trained when the threshold was too small ($<5\%$), but there was negligible performance impact for threshold [5\%, 30\%]. Thus, we chose the threshold as 10\% of maximum intensity for all subsequent studies.

The GRAPPA reconstruction time was approximately 30 s for multi-channel data using the settings detailed above, and approximately 10 min for four channels and 2 min for single channel data. In contrast, the proposed network required less than 41 ms for a multi-channel image and approximately 30 ms for a single channel image. Phase reconstruction took approximately 61 ms for each coil, since the magnitude network must be run first to obtain the phase mask. Since individual coil image reconstructions can be computed in parallel, the total reconstruction time was also approximately 61 ms. Even when  the coil images were sequentially reconstructed, total time for four phase images was less than 250 ms. Reconstruction time using 15 KM iterations was approximately 9 s, which is still significantly shorter than for the current existing CS reconstruction algorithms.

\section{Discussion}

Since the network was implemented in the image domain, different artifact patterns affect the algorithm performance. Some scan parameters, such as FOV or acquisition matrix size, can also affect the artifact pattern, and MR images have widely different contrast and orientation. 
One way to address artifact pattern differences is to incorporate images with many different acquisition parameters into the training data set. For example, the training data use in the current study included images from spin echo and GRE sequences with various TEs and TRs, which helped make the algorithm more robust.  The multi-scale network using dual-frame U-net was more robust than the single resolution network  due to its large receptive field that could capture diverse artifact patterns.
%
%
%
%
%

Phase wrapping could possibly cause problems with network training, which could require using phase unwrapping algorithms.  However, no problems occurred in these experiments, which did not use phase unwrapping, because the network was trained with wrapped phase images, hence the final complex images were not affected by phase wrapping. Although the phase masks meant some pixels within the ROI were set to zero, this had no apparent effect on final reconstruction, since they were isolated noise. This arises from robust neural network training, which uses a variety of training sets. If some phases were reconstructed to masked region pixels, they would have no effect on the final complex valued reconstruction, since the magnitudes of those pixels were close to zero.

%
%

The proposed method provided  the best reconstruction results among the other CS and parallel imaging algorithms compared.  The KM method utilizes k-space measurement, which also improved reconstruction performance. However, there was still some blur in the reconstruction images, believed to stem from the $l_2$ loss function employed for network training. 
However, it is difficult to  claim that convolution operation can cause  blurring,  since many super-resolution algorithms use CNNs~\cite{dong2014learning,kim2016accurate,ledig2016photo}.  
Note that the $l_2$ function is good for reducing MSE errors, but not for perceptual image quality, since it is minimized by averaging all possible outputs~\cite{zhao2015l2,  pathak2016context, zhang2016colorful}.  Thus, a different loss function, such as a generative adversarial network~\cite{goodfellow2014generative}, could be used to improve visual quality~\cite{mardani2017deep}.   For example, deep learning based X-ray CT reconstruction quality is significantly improved using adversarial and perceptual loss~\cite{yang2017low,wolterink2017generative}. Thus, we believe  that employing a different  loss function is a promising research direction, and this will be investigated in future studies.

%
%

\section{Conclusion}\label{sec:conclusion}

This paper proposed  a deep residual learning network for reconstruction of MR images from accelerated MR acquisition inspired by recent deep convolutional framelets theory. The proposed network was designed  to learn global artifact patterns using a multi-resolution network architecture. Network training took several hours, but reconstruction could be quickly performed once trained. Very short reconstruction time from the proposed network was a major advantage compared to CS based iterative reconstruction methods. 

The proposed method operated on not only multi-channel data, but also single-channel data. Even with strong coherent aliasing artifacts, the proposed network successfully learned and removed the aliasing artifacts, whereas current parallel and CS reconstruction methods were unable to remove these artifacts.  Significant advantages of both computational time and reconstruction quality show the proposed deep residual learning approach is a promising research direction for accelerated MRI and has great potential impact.

\section*{Acknowledgement}

This study was supported by the Korea Science and Engineering Foundation (grants NRF-2016R1A2B3008104 and NRF-2013M3A9B2076548), National Institutes of Health Blueprint Initiative for Neuroscience Research (grant U01MH093765), National Institutes of Health (grant P41EB015896), and National Institutes of Health NIBIB (grant K99/R00EB012107). Pre-training data were provided by the Human Connectome Project, MGH-USC Consortium.

\appendix
\numberwithin{equation}{section}

\subsection{Deep Convolutional Framelets: A Review}
\label{ap:framelet}

To make this paper self-contained, we briefly review the theory of deep convolutional framelets~\cite{ye2017deep}, which was the main motivation for \eqref{eq:update2}.
%
For simplicity, we explain the theory using the 1D signal model, but the extension to 2D images are straight-forward.

Let $Z=[z_1,\cdots, z_p] \in \Rd^{n\times p}$ denote a  $p$ channel signal with length $n$. We want to find a  signal representation of $Z$ such that  optimal shrinkage behavior can be learned from the training data.  Therefore, we define the non-local basis $\Phi \in \Rd^{n\times n}$, and local basis $\Psi\in \Rd^{pd\times q}$ \cite{ye2017deep} as
$$\Phi=[\phi_1\cdots \phi_n] \in \Rd^{n\times n}, \quad \Psi=[\psi_1\cdots\psi_q]\in\Rd^{pd\times q}$$
and their duals:
$$\tilde\Phi=[\tilde\phi_1\cdots \tilde\phi_n] \in \Rd^{n\times n}, \quad \tilde\Psi=[\tilde\psi_1\cdots\tilde\psi_q]\in\Rd^{pd\times q} \  .$$
%
%
If these bases satisfy the frame condition~\cite{ye2017deep} 
\begin{eqnarray}
 \tilde \Phi \Phi^\top = &\sum_{i=1}^n \tilde\phi\phi^\top &= I_{n\times n},  \label{eq:idPhi} \\
 \Psi \tilde \Psi^{\top} =& \sum_{j=1}^q \psi_j\tilde\psi_j^\top &= I_{pd\times pd}, \label{eq:id}
 \end{eqnarray}
where $I_{n\times n}$ denotes the $n\times n$ identity matrix, then the signal can be expressed as
%
%
\begin{eqnarray} \label{eq:Zout}
Z 
&=& \left(\Phi C\right) \circledast \nu(\tilde \Psi) , 
\end{eqnarray}
where 
\begin{eqnarray}\label{eq:coef}
C &=&  \Phi^\top \left( Z \circledast  \overline\Psi\right) \quad \in \Rd^{n\times q} 
 \end{eqnarray}
is the framelet coefficient matrix, with the filters
\begin{eqnarray}\label{eq:tauZ}
\Psi &=&\begin{bmatrix}   \overline\psi_1^1 & \cdots &   \overline\psi_q^1  \\ \vdots & \ddots & \vdots \\
\overline\psi_1^p & \cdots &  \overline\psi_q^p 
\end{bmatrix}  \in \Rd^{dp \times q} \\
\nu(\tilde\Psi) &=&  \frac{1}{d} \begin{bmatrix}  \tilde \psi_1^1 & \cdots &  \tilde \psi_1^p  \\ \vdots & \ddots & \vdots \\
\tilde \psi_q^1 & \cdots &  \tilde \psi_q^p 
\end{bmatrix}  \in \Rd^{dq \times p},
\end{eqnarray}
and $\overline \psi_i$ denotes the index reversed vector of $\psi_i$. The convolutions in \eqref{eq:Zout}  and  \eqref{eq:coef} correspond to standard multi-channel convolutions  in the CNN~\cite{ye2017deep}.

The non-local basis $\Phi$ corresponds to the generalized pooling operation, whereas the local basis $\Psi \in \Rd^{pd\times q}$ is a $p$ input, $q$ output multichannel convolutional filter with length $d$~\cite{ye2017deep}. If $q$ is not sufficiently large to satisfy the frame condition \eqref{eq:id}, the signal decomposition corresponds to the low rank Hankel matrix approximation~\cite{ye2017deep}.

This simple convolutional framelet expansion using   \eqref{eq:Zout} and \eqref{eq:coef} is so powerful that the deep CNN with encoder-decoder architecture emerges from them.  
The $L$-layer implementation of the encoder-decoder architecture is defined recursively as
 \begin{eqnarray*}
\tilde\Qc\left( Z;  \{\Psi^{(j)},\tilde \Psi^{(j)}\}_{j=1}^L\right) 
&=&  \left( \tilde\Phi^{(1)}\hat C^{(1)} \right)\circledast \nu(\tilde \Psi^{(1)}) 
\end{eqnarray*}
where $\hat C^{(i)}$ is calculated from decoder as
\begin{eqnarray*}
\hat C^{(i)}  &=&\begin{cases} 
\left(\tilde\Phi^{(i+1)} \hat C^{(i+1)} \right) \circledast \nu (\tilde \Psi^{(i+1)}),  & 1\leq i <L \\
C^{(L)}, & i=L \end{cases},
\end{eqnarray*}
and the encoder part expansion coefficients are 
\begin{eqnarray*}
C^{(i)}  &=&\begin{cases} \Phi^{(i)\top} \left(C^{(i-1)} \circledast \overline\Psi^{(i)}\right),  & 1\leq i \leq L \\
Z, & i=0 \end{cases}
\end{eqnarray*}

In network training, non-local bases are chosen a priori to satisfy the frame condition \eqref{eq:id}, whereas local bases $\{\Psi^{(j)},\tilde \Psi^{(j)}\}_{j=1}^L$ are learned from the training data. Thus, the network can be trained to have a low dimensional manifold structure by controlling the number of filter channels~\cite{ye2017deep}.
%
%
%
Extending this observation, composite mapping $\tilde\Qc$ from magnitude and phase learning can be constructed as
\begin{eqnarray*}
\tilde \Qc(\Xc) =  {\Qc_\Mc}\left(|\Xc|,  \{\Psi_\Mc^{(j)},\tilde \Psi_\Mc^{(j)}\}\right)e^{j {\Qc_\Pc}(\angle \Xc;  \{\Psi_\Pc^{(j)},\tilde \Psi_\Pc^{(j)}\})},
\end{eqnarray*}
where $\{\Psi_\Mc^{(j)},\tilde \Psi_\Mc^{(j)}\}_{j=1}^L$ and $\{\Psi_\Pc^{(j)},\tilde \Psi_\Pc^{(j)}\}_{j=1}^L$ are convolution filters for the magnitude and phase networks, respectively.

\subsection{ Dual frame U-Net Derivation} \label{ap:dual_unet}

For deep convolutional framelets we are interested in   non-local bases, $\Phi$, and their duals, $\tilde \Phi$, that satisfy the frame condition in \eqref{eq:idPhi}. 
This greatly simplifies network design, since network shrinkage can be controlled using the trainable  local bases.
However, the non-local basis for U-net   with average pooling, as shown in Fig.~\ref{fig:Unet}(a), 
\begin{eqnarray}
\Phi=  \frac{1}{\sqrt{2}}\begin{bmatrix} 1 & 0 & \cdots & 0 \\ 1 & 0 & \cdots & 0 \\ 
  \vdots & \vdots & \ddots & \vdots \\ 0 & 0 & \cdots & 1  \\ 0 & 0 & \vdots & 1 \end{bmatrix} , \label{eq:apool}
 \end{eqnarray}
does not satisfy the frame condition \eqref{eq:idPhi}, resulting in high frequency signal loss~\cite{ye2017deep}. A bypass connection (see Fig.~\ref{fig:Unet}(a)) is usually used in U-Net to overcome this limitation, but the bypass does not completely solve the problem, since low frequency content is duplicated~\cite{ye2017deep}.

Therefore, we employ the dual-frame U-net~\cite{han2017framing} to overcome this limitation. The bypass connection mean convolutional framelet coefficients in \eqref{eq:coef} should be replaced with the augmented framelet coefficient matrix,
\begin{eqnarray}\label{eq:Y}
C_{aug}  &=& \Phi_{aug}^\top  (f\circledast \overline \Psi)   = \begin{bmatrix} S \\ C \end{bmatrix},
\end{eqnarray}
where 
\begin{eqnarray}\label{eq:A}
\Phi_{aug}= \begin{bmatrix} I    & \Phi \end{bmatrix} , ~ S= f\circledast \overline \Psi,~  C =  \Phi^\top (f\circledast \overline \Psi) . 
\end{eqnarray}
Our goal is to explicitly derive the dual-frame $\tilde\Phi_{aug}$  that satisfies the frame condition \eqref{eq:idPhi}. From the matrix identity $(\Phi_{aug}^\top \Phi_{aug})^{-1}\Phi_{aug}^\top \Phi_{aug}=I$, the dual frame   is 
\begin{eqnarray}
\tilde \Phi_{aug} &=&  (\Phi_{aug}^\top \Phi_{aug})^{-1}\Phi_{aug}^\top  \notag\\
&=&  ( I + \Phi \Phi^\top)^{-1} \begin{bmatrix} I  & \Phi \end{bmatrix}  \notag\\
&=&  \left(I -\Phi \Phi^\top/2 \right) \begin{bmatrix} I  & \Phi \end{bmatrix} \notag \\
&=& \begin{bmatrix} I-\Phi\Phi^\top/2 & \Phi/2 \end{bmatrix},
\end{eqnarray}
where we use the matrix inversion lemma and orthogonality $\Phi^\top\Phi =I$ for average pooling in \eqref{eq:apool} for the last equality.
Using the  framelet coefficients $C_{aug}$ in \eqref{eq:Y},  the reconstructed signal using the dual-frame is 
\begin{eqnarray}
\hat Z  &=& \tilde \Phi_{aug} C_{aug} \\
&= & \left( I-\frac{\Phi\Phi^\top}{2}\right) S + \frac{1}{2} \Phi C \notag \\
&=& S + \frac{1}{2} \underbrace{\Phi}_{\mbox{unpooling}} \overbrace{(C - \Phi^\top S)}^{\mbox{residual}}. \label{eq:dualUnet}
\end{eqnarray}
Equation~\eqref{eq:dualUnet} suggests a network structure for the dual frame U-Net. In contrast to U-net,  the residual signal at low resolution is up-sampled through the unpooling layer. This can be implemented using an additional bypass connection for the low resolution signal as shown in   Fig.~\ref{fig:Unet}(b).

\subsection{Application to 7T Data}

\begin{figure}[!hbt]    
\center{\includegraphics[width=9cm]{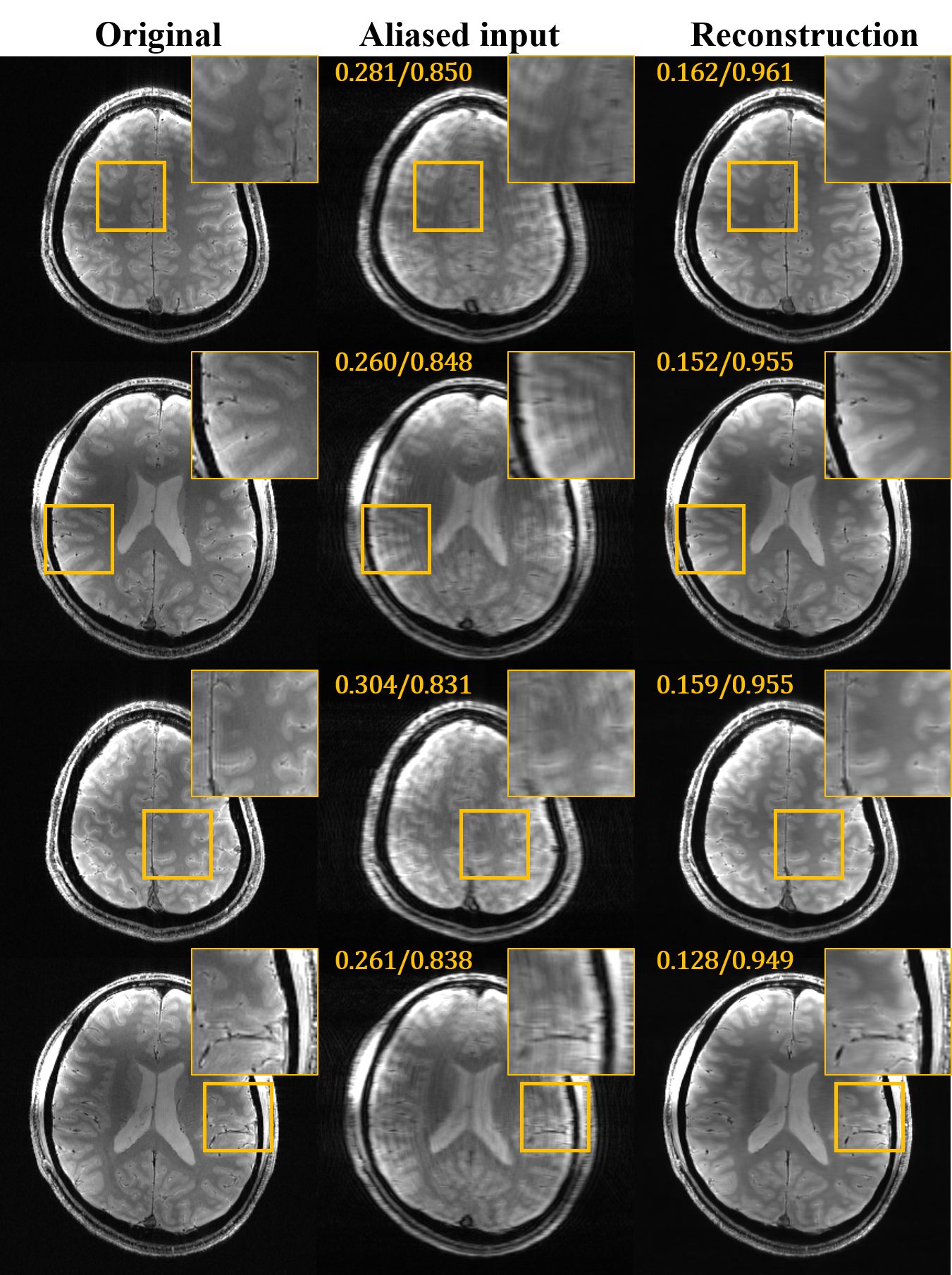}}
\caption{Magnitude reconstruction for 3 test data slices from 7T MR images. Uniform random sampling with 20\% ACS lines as shown in Fig.~\ref{fig:SamplingRecon}(e).}
\label{fig:Result7T}
\end{figure}

We applied the propose algorithm to 7T images to demonstrate it applicability for high resolution images, as shown in Fig.~\ref{fig:Result7T}.  The 7T axial brain images were acquired in Cartesian coordinates from a 7T MR scanner (Philips, Achieva). A total of 567 images were scanned from nine subjects (63 slices per subject), with TR = 831 ms, TE = 5 ms, slice thickness = 0.75 mm, 292$\times$ 292 acquisition matrix, 32 head coils, FOV = 240$\times$ 240 mm, FA = 15$^\circ$, and lateral to lateral PE direction.   Images from 7 subjects were used for training (441 images),  1 subject for validation (63 images), and  the final subject for testing (63 images).

As with the previous studies above, we used uniform random sampling with 20\% ACS lines, and the proposed algorithm significantly improved image quality and reduced aliasing artifacts.

\end{document}